\documentclass{article}
\usepackage{ijcai}

\usepackage{times}
\usepackage{soul}
\usepackage{url}
\usepackage[hidelinks]{hyperref}
\usepackage[utf8]{inputenc}
\usepackage[small]{caption}
\usepackage{graphicx}
\usepackage{amsmath}
\usepackage{amsthm}
\usepackage{booktabs}
\usepackage[switch]{lineno}
\usepackage{amssymb}
\usepackage{amsmath}
\usepackage{tkz-graph}
\usepackage[procnumbered,ruled,linesnumbered,vlined ]{algorithm2e} 
\SetKwInput{KwInput}{Inputs}
\SetKwInput{KwOutput}{Output}

\title{Resource Constrained Pathfinding with A* and Negative Weights}

\author{
 Saman Ahmadi\textsuperscript{*\rm 1},
 Andrea Raith\textsuperscript{\rm 2},
    Mahdi Jalili\textsuperscript{\rm 1}
\affiliations
\textsuperscript{\rm 1} School of Engineering, RMIT University, Australia\\
\textsuperscript{\rm 2} Department of Engineering Science, University of Auckland, New Zealand
\emails
\ *Saman.Ahmadi@rmit.edu.au
}

\begin{document}

\maketitle

\begin{abstract}
Constrained pathfinding is a well-studied, yet challenging network optimisation problem that can be seen in a broad range of real-world applications.
Pathfinding with multiple resource limits, which is known as the Resource Constrained Shortest Path Problem (RCSP), aims to plan a cost-optimum path subject to limited usage of resources.
Given the recent advances in constrained and multi-criteria search with A*, this paper introduces a new resource constrained search framework on the basis of A* to tackle RCSP in large networks, even in the presence of negative cost and negative resources.
We empirically evaluate our new algorithm on a set of large instances and show up to two orders of magnitude faster performance compared to state-of-the-art RCSP algorithms in the literature.
\end{abstract}

\section{Introduction}
The Resource Constrained Shortest Path problem (RCSP) is a classic yet challenging network optimisation problem. 
The objective in point-to-point RCSP is finding a cost optimum path between two locations of the network with limited usage of (one or multiple) resources. 
Many real-word problems can be modelled as an RCSP instance. 
Examples are planning time-constrained explorable non-shortest routes in roaming navigation applications \cite{OtakiMYS23}, and finding shortest paths with battery constraints for electric vehicles \cite{AhmadiTHK21_cp}.
As a subproblem, RCSP has been utilised to solve orienteering problems \cite{VansteenwegenSO11}, or the vehicle routing problem \cite{FeilletDGG04}.

RCSP and its single constraint variant, known as Weight Constrained Shortest Path (WCSP), have been studied in the literature for decades. 
Traditional solutions to RCSP are generally based on path ranking, dynamic programming (labeling) or branch-and-bound (B\&B) approaches \cite{pugliese2013survey,feroneffp20}.
Recent RCSP solutions, however, have found the labeling method more effective in solving large RCSP instances.
The A*-based algorithm of \citeauthor{thomas2019exact}~\shortcite{thomas2019exact}, known as \textsf{RCBDA*}, utilises the dynamic programming approach of \citeauthor{righini2006symmetry}~\shortcite{righini2006symmetry} to conduct bidirectional A* search with only half of the resources available to each direction.
Complete paths in this framework are obtained by joining forward and backward partial paths.
\citeauthor{thomas2019exact} show that {\textsf{RCBDA*} can effectively outperform the B\&B method of \citeauthor{lozano2013exact}~\shortcite{lozano2013exact} and the path ranking method of \citeauthor{sedeno2015enhanced}~\shortcite{sedeno2015enhanced} on large-size instances.
\citeauthor{cabrera2020exact}~\shortcite{cabrera2020exact} later developed a parallel bidirectional search framework on the basis of the B\&B method of \citeauthor{lozano2013exact}~\shortcite{lozano2013exact}, known as \textsf{Pulse}. 
Their \textsf{BiPulse} algorithm limits the depth of the \textsf{Pulse} search to delay the process of deep branches with the help of the queuing strategy of \citeauthor{BolivarLM14}~\shortcite{BolivarLM14}.
\citeauthor{cabrera2020exact} reported faster performance and more solved cases with \textsf{BiPulse} compared to \textsf{RCBDA*} on medium-size RCSP instances.
\textsf{WCBA*}\cite{AhmadiTHK22_socs} and \textsf{WCA*} \cite{AhmadiTHK23_Networks} are two other A*-based methods designed to tackle WCSP, both outperforming \textsf{RCBDA*} and \textsf{BiPulse} on large graphs, but for instances with one resource only. 
Quite recently, \citeauthor{ren2023erca}~\shortcite{ren2023erca} have developed an A*-based RCSP method, called \textsf{ERCA*}, that takes advantage of the recent advancements in multi-objective search with A* \cite{ulloa2020simple,PulidoMP15,RenZRLC22} to improve the efficiency of pruning rules, in particular, through lazy dominance checks of partial paths using balanced binary search trees.
\citeauthor{ren2023erca} reported several orders of magnitude faster runtime for \textsf{ERCA*} when compared with \textsf{BiPulse}.
However, the performance of \textsf{ERCA*} against \textsf{RCBDA*} remains unknown.

\textbf{RCSP with negative weights}:
Many real-world RCSP problems deal with attributes that are negative in nature, such as energy recuperation in electric vehicles, or attributes that may exhibit negative values in specific circumstances, such as negative reduced costs in column generation or prize-collecting problems.
None of the recent RCSP solutions are designed to deal with problems containing negative edge attributes.
Among the existing solutions, the labeling method of \citeauthor{FeilletDGG04}~\shortcite{FeilletDGG04} is one of the first attempts to solve the elementary variant of RCSP with negative costs and cycles.
\citeauthor{righini2008new}~\shortcite{righini2008new} and \citeauthor{PuglieseG12}~\shortcite{PuglieseG12} also designed dynamic programming approaches in the column generation context. More recent work have found B\&B \cite{LozanoDM16} and mixed integer programming \cite{BuiDP16} also effective in solving RCSP with negative costs, mainly in vehicle routing applications. 
However, solutions to the elementary variant of RCSP are either not practical for large RCSP problems, or designed for negative costs only.

This paper introduces \textsf{NWRCA*}, a novel search procedure for exact Resource Constrained pathfinding using A* in problems with Negative edge Weights, but no negative cycles.
\textsf{NWRCA*} leverages the best-first search strategy of the recent A*-based constrained and multi-objective search algorithms to improve the efficiency of exhaustive search of A*.
We show through extensive experiments the success of \textsf{NWRCA*} in solving large RCSP instances, where it outperforms the state of the art by up to two orders of magnitude.

\section{Problem Definition}
Consider an RCSP problem with $\mathit{d}$ resource limits $\{R_1, R_2, \dots, R_d\}$ is provided as directed graph $G=(S,E)$ with a finite set of states $S$ and a set of edges $E \subseteq S \times S$.
Every edge $e \in E$ of the graph has $\mathit{d}+1 \in \mathbb{N}$ (potentially negative) attributes, represented as $(\mathit{cost}, \mathit{resource}_1, \dots, \mathit{resource_{d}})$, accessible via the cost function ${\bf cost}:E \to \mathbb{R}^{d+1}$.
Every boldface function in our notation returns a vector.
For the sake of simplicity in our algorithmic description, we consider each resource as a form of non-primary cost, so we have ${\textbf{cost}}=(\mathit{cost}_1, \mathit{cost_2}, \dots, \mathit{cost_{d+1}})$ as cost vector, with $\mathit{cost}_1$ representing the primary cost of the problem and ($\mathit{cost}_2, \dots, \mathit{cost_{d+1}})$ denoting the resource usage vector.
A path is a sequence of states $u_i \in S$ with $i \in \{1, \dots, n \}$ and $(u_i,u_{i+1}) \in E$ for $i \in \{1\dots n-1\}$.
The $\textbf{cost}$ vector of path $\pi=\{u_1,u_2,u_3,\dots,u_n\}$ is then the sum of corresponding attributes on all of the edges constituting the path, namely $\textbf{cost}(\pi) = \sum_{i=1}^{n-1}{\textbf{cost}(u_i,u_{i+1})}$.
Since costs can be negative values, we say path $\pi$ forms an elementary negative cycle on $\mathit{cost_k}$ if 
i) $\mathit{cost}_k(\pi) < 0$; 
ii) $u_i \neq u_j$ for any $i,j \in \{1\dots n-1, i \neq j\}$; 
iii) $u_1=u_n$. 

The point-to-point RCSP problem aims to find a set of non-dominated $\mathit{cost}_1$-optimal paths between $\mathit{start} \in S$ and $\mathit{goal} \in S$ such that the resource usages of the paths are within the given limits, i.e., for every solution path $\pi^*$, we must have $\mathit{cost}_{k+1}(\pi^*) \leq R_{k}$  for every resource $k \in \{1,\dots, d\}$.

Following the conventional notation in the heuristic search literature, we define our search objects to be \textit{nodes} (equivalent to partial paths from the $\mathit{start}$ state).
A node $x$ is a tuple that contains the main information of the partial path to state $s(x) \in S$,  where $s(x)$ is a function returning the state associated with $x$. 
Node $x$ traditionally stores a value pair $\textbf{g}(x)$ which measures the $\mathit{\bf cost}$ of a concrete path from the $\mathit{start}$  state to $s(x)$.
In addition, node $x$ is associated with a value pair $\textbf{f}(x)$, which estimates the $\mathit{\bf cost}$ of a complete $\mathit{start}$-$\mathit{goal}$  path via $x$; and also a reference $\mathit{parent}(x)$ which indicates the parent node of $x$.
Further, the operator $\mathrm{Tr}(\mathbf{v})$ truncates the primary (first) cost of the vector $\mathbf{v}$. 
For example, $\left({g_2}(x), \dots, {g_{d+1}}(x)\right)$ is the truncated vector of ${\bf g}(x)$.

We consider all operations of the cost vectors to be done element-wise.
For example, we define ${\bf g}(x) + {\bf g}(y)$ as $\left({g_1}(x) + {g_1}(y), \dots, {g_{d+1}}(x) + {g_{d+1}}(y)\right)$.
We use $\preceq$ or $\leq_{lex}$ symbols in direct comparisons of cost vectors, e.g. ${\bf g}(x) \preceq {\bf g}(y)$ denotes
${g_k}(x) \leq {g_k}(y)$ for all $k \in \{1,\dots,d+1\}$ and ${\bf g}(x) \leq_{lex} {\bf g}(y)$ means the cost vector ${\bf g}(x)$ is lexicographically smaller than or equal to ${\bf g}(y)$.
Analogously, we use $\nprec$ or $\nleq_{lex}$ symbols if the relations cannot be satisfied.

\noindent \textbf{Definition\ }
Node $\mathit{y}$ is weakly dominated by node $\mathit{x}$ if we have ${\bf g}(x) \preceq {\bf g}(y)$; $\mathit{y}$ is dominated by $\mathit{x}$ if ${\bf f}(x) \preceq {\bf f}(y)$ and ${\bf f}(x) \neq {\bf f}(y)$; $y$ is non-dominated if ${\bf g}(x) \npreceq {\bf g}(y)$.

The main search in A*-based solution methods is guided by $\mathit{start}$-$\mathit{goal}$ cost estimates or {\bf f}-values, which are traditionally established based on a heuristic function
${\bf h}: S \rightarrow \mathbb{R}^{d+1}$ 
\cite{hart1968formal}.
In other words, for every search node $x$, we have ${\bf f}(x)={\bf g}(x)+{\bf h}(s(x))$ where ${\bf h}(s(x))$ estimates lower bounds on the ${\bf cost}$ of paths from $s(x)$ to the $\mathit{goal}$ state.

\noindent \textbf{Definition} 
For each $k \in \{1,\dots, d+1\}$, we say $h_k: S \rightarrow \mathbb{R}$ is admissible iff $h_k(u) \le \mathit{cost}_k(\lambda^*)$ for every $u \in S$ where $\lambda^*$ is the optimal path on $\mathit{cost}_k$ from state $\mathit{u}$ to the $\mathit{goal}$ state. 
It is also consistent if we have ${ h_k}(u) \le {cost}_k(u,v) + {h_k}(v)$ for every edge $(u,v) \in E$. 

\section{Resource Constrained Pathfinding with A*}
Recent solutions to constrained and multi-objective pathfinding have found A* an effective tool in better guiding the search towards optimal solutions \cite{AhmadiSHJ24,ren2023erca,LTOMOA3,TMDA23}.
Constrained pathfinding with A* involves a systematic search by \textit{expanding} nodes in best-first order.
That is, the search is led by a partial path that shows the lowest $\mathit{\bf cost}$ estimate or ${\bf f}$-value.
Each iteration involves three main steps: \\
i) Extraction: remove one least-$\mathit{\bf cost}$ node from a priority queue, known as $\mathit{Open}$ list.\\
ii) Dominance check: ensure the extracted node is not dominated by previous expansions in terms of $\mathit{\bf cost}$ or ${\bf g}$-values; \\
iii) Expansion: generate new descendant nodes, check them for dominance and bounds (resource limits), and, if not pruned, store them in $\mathit{Open}$ for further expansion.\\
The search in this framework generally terminates once a solution is found, or there is no node in $\mathit{Open}$ to explore, where the latter means there is no feasible solution.
If there is only one resource, it is shown that dominance pruning can be done in constant time when performing best-first search \cite{AhmadiTHK22_socs,AhmadiTHK23_Networks}.
Nonetheless, this would not be the case in RCSP with multiple resources and the dominance pruning is an expensive task.
With this introduction, we now describe our new A*-based RCSP search framework \textsf{NWRCA*}.
Our new approach is equipped with efficient pruning rules, finds all non-dominated optimum solutions, and more importantly, handles negative edge attributes.
\subsection{\textsf{NWRCA*}'s High-Level Description:}
Algorithm~\ref{alg:high-level} provides a high-level description of \textsf{NWRCA*}, presenting it as a merged procedure for establishing heuristic functions and detecting negative cycles.
\textsf{NWRCA*} borrows some of its features from our multi-objective search algorithm \textsf{NWMOA*} \cite{AhmadiSHJ24}.
It first initialises a global upper bound ${\bf \bar{f}}$. 
The upper bound on the primary cost is unknown and is set to a large value, whereas the upper bound on each resource is set to its given budget. 
The global upper bound will be used in the main search to prune paths violating the resource limits. 
It then initialises the ${\bf h}$-value of all states with a vector of $d+1$ large values, followed by eliminating all states not reachable from $\mathit{start}$ (e.g., using breath-first search) to form a reduced graph.
Nodes not reachable from source cannot be part of any solution, and thus can be safely removed.
Like other A*-based algorithms, \textsf{NWRCA*} requires establishing its heuristic function prior to running the main search.
The algorithm then conducts for each $k \in \{1, \dots, d+1\}$ a backward one-to-all single-objective search to compute $\mathit{cost}_k$-optimal paths from $\mathit{goal}$.
When the graph is negative-cycle free on $\mathit{cost}_k$, $h_k$ becomes a perfect heuristic function, meaning that all lower bounds are bounded and $\mathit{cost}_k$-optimal.
Finding lower bounds can be as simple as $d+1$ runs of the Bellman-Ford algorithm \cite{bellman1958routing,ford1956network}, or the Dijkstra's algorithm with re-expansions allowed \cite{johnson1973note}.
However, to ensure the heuristic function is consistent, we need all $\mathit{start}$-$\mathit{goal}$ paths be negative-cycle free.
The condition $h_k(u)=-\infty$ at line~\ref{alg:high-level:cycle} of Algorithm~\ref{alg:high-level} denotes $u$ belongs to a negative cycle, and thus there is a negative cycle on a $\mathit{start}$-$\mathit{goal}$ path (via $u$).
With both lower and upper bounds established, we can proceed with the \textsf{NWRCA*}'s lower-level search (via line~\ref{alg:high-level:mult-search}).

\begin{algorithm}[t]
\small
\caption{\textsf{NWRCA*} High Level}
\label{alg:high-level}
\DontPrintSemicolon
\KwIn{An RCSP Problem ($\mathit{G}$, $\mathit{\mathit{start}}$, $\mathit{\mathit{goal}}, R_1, \dots, R_d$)}

\KwOutput{A non-dominated $\mathit{cost}_1$-optimal solution set}
$ \overline{\bf f} \gets (\infty, R_1, \dots, R_d)$ \;
$ {\bf h}(u) \gets {\bf \infty} \ \forall u \in S$ \;
$S \gets$ Remove from $S$ states not reachable form $\mathit{start}$\;
        
    \For{$ k \in \{1, \dots, d + 1 \}$}
        {  
        \ForEach{ $u \in S$}
        {
        $h_k(u) \gets$ $\mathit{cost}_k$-optimal path from $u$ to $\mathit{goal}$\;
        \If {$h_k(\mathit{u}) = -\infty$ \label{alg:high-level:cycle}} 
        {\Return{$\emptyset$}}
        }

        }
$\mathit{Sols} \gets $ Constrained Search on ($\mathit{G}$, {\bf h}, $\mathit{\mathit{start}}$, $\mathit{\mathit{goal}}$, $\overline{\bf f}$) \label{alg:high-level:mult-search}\;    
\Return{$\mathit{Sols}$}
\end{algorithm}

\subsection{\textsf{NWRCA*}'s Low-level Search}
\begin{algorithm}[!t]
\small
\caption{Constrained Search of \textsf{NWRCA*} }
\label{alg:nwrca}
\DontPrintSemicolon
 \KwIn{An RCSP Problem ($\mathit{G}$, {\bf h}, $\mathit{\mathit{start}}$, $\mathit{\mathit{goal}}$, $\overline{\bf f}$)}
 
 \KwOutput{A non-dominated $\mathit{cost}_1$-optimal solution set}
 
 $\mathit{Open} \gets \emptyset $ , $\mathit{Sols} \gets \emptyset $\;
$ \mathrm{G_{cl}^{Tr}}(u) \gets \emptyset$ \ $\forall u \in S$\ \;
$ \mathrm{{\bf g}_{last}^{Tr}}(u) \gets {\bf \infty}$ \ $\forall u \in S$\ \label{alg:nwrca:init_last}\;

 $x \gets $ new node with $s(x) = \mathit{start}$\ \;
 $ {\bf g}(x) \gets {\bf 0}$ ,
 $ {\bf f}(x) \gets {\bf h}(\mathit{start})$ ,
 $parent(x) \gets \mathit{null} $ \;
Add $x$ to $Open$\;


\While{$Open \neq \emptyset$ \label{alg:nwrca:iter}}
{
 
  {Extract from $Open$ node $x$ with the smallest $f_1$-value \label{alg:nwrca:least_cost} \;}
       
      \lIf{ $f_1(x) > \overline{f_1}$ \label{alg:nwrca:terminate}} 
      {\textbf{break} }
      
      { \lIf{ $\mathrm{{\bf g}_{last}^{Tr}}(s(x)) \preceq \mathrm{Tr}({\bf g}(x))$ \label{alg:nwrca:qdom1}} 
      {\textbf{continue} } 
      }
      
       \If{ $\mathtt{IsDominated}(\mathrm{Tr}({\bf g}(x)),\mathrm{G_{cl}^{Tr}}(s(x)))$ \label{alg:nwrca:dom}} 
      {\textbf{continue}}
 
      
      $\mathtt{Consolidate}(\mathrm{Tr}({\bf g}(x)),\mathrm{G_{cl}^{Tr}})$ \label{alg:nwrca:consolidate}\;
     
    $\mathrm{{\bf g}_{last}^{Tr}}(s(x)) \gets \mathrm{Tr}({\bf g}(x))$ \label{alg:nwrca:last_path}\;

\If{$ s(x) = \mathit{goal} $ \label{alg:nwrca:goal} \label{alg:nwrca:sol0}}
    {      
        
        $\overline{f_1} \gets f_1(x) $\;
        $i \gets$ $|Sols|$ \label{alg:nwrca:sol1} \;
        \While{$i >= 1$}
        {
        $z \gets$ Node at index $i$ of $Sols$ \;
        \lIf {${\bf g}(x) \preceq {\bf g}(z)$}
        {remove $z$ from $Sols$}
        $i \gets (i - 1)$ \label{alg:nwrca:sol2}\;
        }
        
        Add $x$ to the end of $Sols$ \label{alg:nwrca:sol3}\;
        {\bf continue \label{alg:nwrca:sol4}} \;
    }
    
    \ForEach{$t \in Succ(s(x))$}
        {  $y \gets $ new node with $s(y) = t$ \label{alg:nwrca:expansion1}\; 
             ${\bf g}(y) \gets {\bf g}(x) + {\bf cost} (s(x),t)$ \; 
             ${\bf f}(y) \gets {\bf g}(y) + {\bf h} (t)$ \;
             $parent(y) \gets x$ \label{alg:nwrca:expansion2}\; 
             
             \lIf{ ${\bf f}(y) \npreceq \overline{\bf f}$
       \label{alg:nwrca:bound}} 
      {\textbf{continue} }
      
        \lIf{ $\mathrm{{\bf g}_{last}^{Tr}}(t) \preceq \mathrm{Tr}({\bf g}(y))$ 
       \label{alg:nwrca:qdom2}} 
      {\textbf{continue} } 
            
            
            Add $y$ to $Open$\; \label{alg:nwrca:expansion3}
            
        }

}
\Return{$Sols$}
\end{algorithm}
\begin{algorithm}
\small
\caption{$\mathtt{IsDominated}$}
\label{alg:Isdominated}
\DontPrintSemicolon
\KwIn{A vector $\mathbf{v}$ and a set of vectors $\mathbf{V}$ in lex. order}
 
 \KwOutput{$\mathit{true}$ if $\mathbf{v}$ is weakly dominated, $\mathit{false}$ otherwise}

    \For{$i \in \{1, \dots, |\mathbf{V}|\}$}
        {  
        $\mathbf{v'} \gets $ Cost vector at index $i$ of $\mathbf{V}$ \;
        \If {$\mathbf{v'} \nleq_{lex} \mathbf{v}$ \label{alg:isdom:check1}}
        {\Return{$\mathit{false}$}}
        \If {$\mathbf{v'} \preceq \mathbf{v}$}
        {\Return{$\mathit{true}$}}
        
        }
\Return{$\mathit{false}$}
\end{algorithm}
\begin{algorithm}
\small
\caption{$\mathtt{Consolidate}$}
\label{alg:Consolidate}
\DontPrintSemicolon
\KwIn{A vector $\mathbf{v}$ and a set of vectors $\mathbf{V}$ in lex. order}
 
 \KwOutput{$\mathbf{V}$ updated with dominated cost vectors removed}
 
$\mathit{i} \gets |\mathbf{V}|$ \;
    \While{$\mathit{i} \geq 1 $}
        {  
        $\mathbf{v'} \gets $ Cost vector at index $\mathit{i}$ of $\mathbf{V}$ \;
        \If {$\mathbf{v} \nleq_{lex} \mathbf{v'}$}
        {
        {\bf break}        
        }
            
        \If {$\mathbf{v} \preceq \mathbf{v'}$}
        {
        Remove $\mathbf{v'}$ from $\mathbf{V}$ \label{alg:Consolidate:remove}
        }
        $\mathit{i} \gets (\mathit{i} - 1)$ \;

        }
 Insert $\mathbf{v}$ to index $(\mathit{i}+1)$ of $\mathbf{V}$ \label{alg:Consolidate:insert} \;
 \Return{}
\end{algorithm}

Algorithm~\ref{alg:nwrca} shows the pseudo-code of the constrained search of \textsf{NWRCA*}.
It starts with initialising the solution set $\mathit{Sols}$, and the search's priority queue $\mathit{Open}$.
It then sets up for every state $u \in S$ a list $\mathrm{G_{cl}^{Tr}}(u)$ and a cost vector $\mathrm{{\bf g}_{last}^{Tr}}(u)$. The $\mathrm{G_{cl}^{Tr}}(u)$ list is responsible for storing the (non-dominated) truncated cost-vector of all previous (closed) expansions with $u$.
The cost vector $\mathrm{G_{cl}^{Tr}}(u)$, initialised with large cost values (line \ref{alg:nwrca:init_last}), keeps track of the most recent expansion of $u$.
\textsf{NWRCA*} then initialises a node with the $\mathit{start}$ state and inserts it into $\mathit{Open}$.
Each iteration of the search starts at line~\ref{alg:nwrca:iter}.

Let $\mathit{Open}$ be a non-empty queue.
Following the queuing method of \textsf{NWMOA*}, the algorithm extracts in each iteration a node $x$ with the smallest $f_1$-value (line~\ref{alg:nwrca:least_cost}), meaning that there is no tie-breaking in place (when multiple paths show the same $f_1$-value).
This technique will help constrained A* undertake less queue operations throughout its exhaustive constrained search, as shown in \citeauthor{AhmadiTHK23_Networks}~\shortcite{AhmadiTHK23_Networks}.

\textbf{Early termination:}
\textsf{NWRCA*} explores nodes in non-decreasing order of their $f_1$-value. 
Given $\overline{f_1}$ as the upper bound on the primary cost, the search can terminate early once it extracts a node with $f_1$-value larger than $\overline{f_1}$ (line~\ref{alg:nwrca:terminate}). 
Although $\overline{f_1}$-value is initially unknown, it will get updated as soon as a feasible $\mathit{cost}_1$-optimal solution is discovered. 

\textbf{Dominance rule:}
Let $x$ be a node extracted from $\mathit{Open}$ with ${\bf f}$-value within the global upper bound.
Because the first dimension is always expanded in sorted $f_1$ order, all previous expansions with $s(x)$ must show $g_1$-value no smaller than $g_1(x)$, meaning that $x$ is already weakly dominated by previous expansions in the first dimension. 
Note that all expansions with $s(x)$ use $h_1(s(x))$ as lower bound.
Thus, the dominance test can be done by comparing the truncated cost vector of $x$, i.e., $\mathrm{Tr}({\bf g}(x))$, with that of previous expansions.

\textbf{Quick dominance check:}
Recent expansions are generally more informed and can be seen as strong candidates for dominance check.
\textsf{NWRCA*} quickly checks $x$ for dominance against the last expanded node of the state associated with $x$, i.e., $s(x)$, before attempting dominance check against all potential candidates (line~\ref{alg:nwrca:qdom1}).

\textbf{Dominance check with $\mathtt{IsDominated}$:}
If the extracted node cannot be quickly dominated, \textsf{NWRCA*} takes $\mathrm{Tr}({\bf g}(x))$ to conduct a rigorous dominance check by comparing $x$ against (potentially all) truncated vectors of previous expansions with $s(x)$, as detailed in Algorithm~\ref{alg:Isdominated}.
Truncated cost vectors are stored in a lexicographical order.
Thus, the dominance test of Algorithm~\ref{alg:Isdominated} does not need to traverse the entire $\mathrm{G_{cl}^{Tr}}(s(x))$ list, and can terminate early when it discovers a candidate with a truncated cost vector not lexicographically smaller than that of $x$ (line~\ref{alg:isdom:check1} of Algorithm~\ref{alg:Isdominated}). 

\textbf{Cycle elimination:}
Since we assume there is no negative cycle on any $\mathit{start}$-$\mathit{goal}$ paths, extending a path to one of its already-visited states through a non-negative cycle will produce a ${\bf g}$-value no smaller than that of the first visit. Thus, nodes forming such cycles are pruned via dominance rule.

\textbf{Lexicographical ordering with $\mathtt{Consolidate}$:}
If $x$ is a non-dominated node, its truncated cost vector $\mathrm{Tr}({\bf g}(x))$ must be stored in $\mathrm{G_{cl}^{Tr}}(s(x))$ to be used in future dominance checks with $s(x)$.
However, $\mathrm{Tr}({\bf g}(x))$ may dominate some vectors of $\mathrm{G_{cl}^{Tr}}(s(x))$.
Truncated vectors lexicographically smaller than $\mathrm{Tr}({\bf g}(x))$ cannot be dominated.
Thus, $\mathtt{Consolidate}$, described in Algorithm~\ref{alg:Consolidate}, iterates backward through $\mathrm{G_{cl}^{Tr}}(s(x))$ to remove vectors dominated by the new non-dominated vector (line~\ref{alg:Consolidate:remove}), and stops as soon as it finds $\mathrm{Tr}({\bf g}(x))$ no longer lexicographically smaller than the candidate vector in $\mathrm{Tr}({\bf g}(x))$.
It then inserts $\mathrm{Tr}({\bf g}(x))$ after the last attempted candidate (line~\ref{alg:Consolidate:insert}), ensuring vectors are maintained in lexicographical order.
Once $\mathrm{Tr}({\bf g}(x))$ is captured, \textsf{NWRCA*} stores a copy of it into $\mathrm{{\bf g}_{last}^{Tr}}(s(x))$ as the most recent expansion (line~\ref{alg:nwrca:last_path} of Algorithm~\ref{alg:nwrca}).

\textbf{Capturing solutions:}
Let $x$ be a node associated with $\mathit{goal}$ (line~\ref{alg:nwrca:sol0} of Algorithm~\ref{alg:nwrca}).
The search has found a tentative solution path.
If $\mathit{Sols}$ is empty, we simply capture $x$ as a tentative solution and add it to the solution set (line~\ref{alg:nwrca:sol3}).
Otherwise, 
it is possible for $x$ to dominate some previous solution nodes in $\mathit{Sols}$.
To this end, the search (linearly) iterates backward through $\mathit{Sols}$, checks the new solution $x$ against the previous solutions via lines~\ref{alg:nwrca:sol1}-\ref{alg:nwrca:sol2}, and removes dominated ones. Note that, since new solutions are already tested for dominance, they are always added to $\mathit{Sols}$.

\textbf{Expansion:}
Let $x$ be a non-dominated node other than a solution node.
Expansion of $x$ involves generating new descendant nodes through $s(x)$'s successors.
Once a descendant node is generated, it is immediately checked against the global upper bounds. 
Given ${\bf h}$ as an admissible heuristic function, nodes showing $f_k$-value larger than the upper bound $\overline{f}_k$ cannot lead to a feasible $\mathit{cost}_1$-optimal solution and thus can be pruned, as scripted at line~\ref{alg:nwrca:bound} of Algorithm~\ref{alg:nwrca}.
Consistent with lazy dominance checks in recent RCSP solutions, \textsf{NWRCA*} delays the (full) dominance check of new nodes until they are extracted from $\mathit{Open}$.
Nonetheless, the quick dominance check 
can still be performed (see line~\ref{alg:nwrca:qdom2}).
Thus, each descendant node will be added to $\mathit{Open}$ if it shows cost estimates within $\overline{\bf f}$, and if it is not dominated by the most recent expansion of the successor state. 

Finally, the algorithm returns $Sols$ as a set containing non-dominated $\mathit{cost}_1$-optimal solution paths to a solvable RCSP instance with negative costs but no negative cycles.

\textbf{Example:}
We further elaborate on the key steps of \textsf{NWRCA*} by solving a sample RCSP instance with three edge attributes (two resources), depicted in Figure~\ref{fig:example}.
We assume resource limits are $R_1=R_2=3$, thus \textsf{NWRCA*} initialises the global upper bound  $\overline{\bf f} \gets (\infty,3,3)$.
The states $u_s$ and $u_g$ denote $\mathit{start}$ and $\mathit{goal}$, respectively. 
Although the graph contains negative cycle (shown in red), the problem has a feasible solution as the cycle is not on any $\mathit{start}$-$\mathit{goal}$ path.
For the states reachable from $\mathit{start}$, the triple inside each state denotes ${\bf h}$-value, calculated in the high level of \textsf{NWRCA*} (Algorithm \ref{alg:high-level}).
Note that each $h_k(u_i)$-value represents a $\mathit{cost}_k$-optimal path from $u_i$ to $u_g$ for $k \in \{1,2,3\}$.
We briefly explain all iterations (It.) of the search for the given instance, with the trace of $\mathit{Open}$, $\mathrm{G_{cl}^{Tr}}$ and $\mathit{Sols}$ illustrated in Table~\ref{tab:example}.
For node extractions, we adopt the Last-In, First-Out (LIFO) strategy in the event of ties in $f_1$-values.

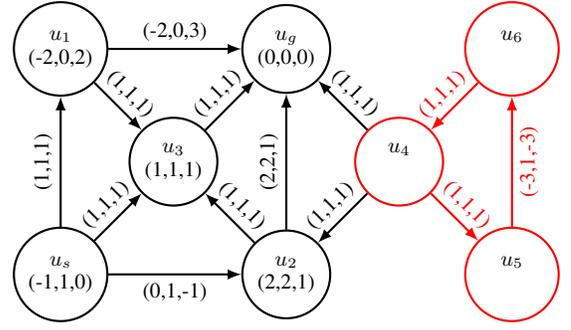
\begin{figure}
\centering
\begin{tikzpicture}[
roundnode/.style={circle, draw=black,  thick, minimum size=5mm},
roundnode2/.style={circle, draw=red,  thick, minimum size=5mm},
scale=0.75, every node/.style={scale=0.90}]

\footnotesize
\node[roundnode,align=center]   at (0, 0) (us)        {$u_s$ \\ (-1,1,0)};
\node[roundnode,align=center]   at (4, 0) (u2)        {$u_2$ \\ (2,2,1)};
\node[roundnode2,align=center]   at (8, 0) (u5)        {$u_5$ \\ \textcolor{white}{(1,0,-2)}};

\node[roundnode,align=center]   at (2, 2) (u3)        {$u_3$ \\ (1,1,1)};
\node[roundnode2,align=center]   at (6, 2) (u4)        {$u_4$ \\ \textcolor{white}{(1,1,0)}};

\node[roundnode,align=center]   at (0, 4) (u1)        {$u_1$ \\ (-2,0,2)};
\node[roundnode,align=center]   at (4, 4) (ug)        {$u_g$ \\ (0,0,0)};
\node[roundnode2,align=center]   at (8, 4) (u6)        {$u_6$ \\ \textcolor{white}{(1,1,-1)}};


\draw[->,-latex,  thick] (us) edge[auto=right] node{(0,1,-1)} (u2);
\draw[->,-latex,  thick] (us) edge[auto=left] node[midway, above,rotate=90]{(1,1,1)} (u1);
\draw[->,-latex,  thick] (us) edge[auto=left] node[midway, above,rotate=45]{(1,1,1)} (u3);
\draw[->,-latex,  thick] (u1) edge[auto=left] node[midway, above,rotate=-45]{(1,1,1)} (u3);
\draw[->,-latex,  thick] (u1) edge[auto=left] node{(-2,0,3)} (ug);
\draw[->,-latex,  thick] (u2) edge[auto=right] node[midway, above,rotate=-45]{(1,1,1)} (u3);
\draw[->,-latex,  thick] (u3) edge[auto=right] node[midway, above,rotate=45]{(1,1,1)} (ug);

\draw[->,-latex,  thick] (u2) edge[auto=right] node[midway, above,rotate=90]{(2,2,1)} (ug);
\draw[->,-latex,  thick, color = red] (u4) edge[auto=left,] node[midway, above,rotate=-45]{(1,1,1)} (u5);
\draw[->,-latex,  thick, color = red] (u5) edge[auto=right] node[midway, below,rotate=90]{(-3,1,-3)} (u6);
\draw[->,-latex,  thick, color = red] (u6) edge[auto=left] node[midway, above,rotate=45]{(1,1,1)} (u4);

\draw[->,-latex,  thick] (u4) edge[auto=right] node[midway, above,rotate=-45]{(1,1,1)} (ug);
\draw[->,-latex,  thick] (u4) edge[auto=left] node[midway, above,rotate=45]{(1,1,1)} (u2);

\end{tikzpicture}

\caption{\small An example graph with three edge attributes and negative cycle (in red). Triples inside the states denote the ${\bf h}$-value.}
\label{fig:example}
\end{figure}
\begin{table}[t]
    \small
    \centering
    \setlength{\tabcolsep}{3.5pt}
    \begin{tabular}{ c   l   c  c }

    \toprule
    It. & $\mathit{Open}$ : [${\bf f}(x), {\bf g}(x), s(x)$] & $\mathrm{G_{cl}^{Tr}}$ & $\mathit{Sols}$\\
    \midrule
    1 & $^{*}x_0$=[(-1,1,0), (0,0,0), $u_{s}$] & $\mathrm{G_{cl}^{Tr}}(u_s)$=[(0,0)] &  \\
   \midrule 
    2 & $^{*}x_1$=[(-1,1,3), (1,1,1), $u_1$] & $\mathrm{G_{cl}^{Tr}}(u_1)$=[(2,1)] &  \\
    & \ \ $x_3$=[(2,2,2), (1,1,1), $u_3$]  &  & \\
    & \ \ $x_2$=[(2,3,0), (0,1,-1), $u_2$] &    & \\
    \midrule
    3 & $^{*}x_3$=[(2,2,2), (1,1,1), $u_3$]  & $\mathrm{G_{cl}^{Tr}}(u_3)$=[(1,1)]  & \\
    & \ \ $x_2$=[(2,3,0), (0,1,-1), $u_2$]  &  & \\
    & \ \ $x_4$=[(3,3,3), (2,2,2), $u_3$] &    & \\
    \midrule
    4 & $^{*}x_6$=[(2,2,2), (2,2,2), $u_g$]  & $\mathrm{G_{cl}^{Tr}}(u_g)$=[(2,2)] & $x_6$\\
    & \ \ $x_2$=[(2,3,0), (0,1,-1), $u_2$]  &  & \\
    & \ \ $x_4$=[(3,3,3), (2,2,2), $u_3$] &    & \\
    \midrule
    5 & $^{*}x_2$=[(2,3,0), (0,1,-1), $u_2$]  & $\mathrm{G_{cl}^{Tr}}(u_2)$=[(1,-1)] & $x_6$\\
    & \ \ $x_4$=[(3,3,3), (2,2,2), $u_3$] &    & \\
    \midrule
    6 & $^{*}x_8$=[(2,3,1), (1,2,0), $u_3$]  & $\mathrm{G_{cl}^{Tr}}(u_3)$=[(1,1),(2,0)] & $x_6$\\
    & \ \ $x_7$=[(2,3,0), (2,3,0), $u_g$]  &  & \\
    & \ \ $x_4$=[(3,3,3), (2,2,2), $u_3$] &    & \\
    \midrule   
    7 & $^{*}x_9$=[(2,3,1), (2,3,1), $u_g$]  & $\mathrm{G_{cl}^{Tr}}(u_g)$=[(2,2),(3,1)] & $x_{6,9}$\\
    & \ \ $x_7$=[(2,3,0), (2,3,0), $u_g$]  &  & \\
    & \ \ $x_4$=[(3,3,3), (2,2,2), $u_3$] &    & \\
    \midrule
    8 & $^{*}x_7$=[(2,3,0), (2,3,0), $u_g$]  & $\mathrm{G_{cl}^{Tr}}(u_g)$=[(2,2),(3,0)] & $x_{6,7}$\\
    & \ \ $x_4$=[(3,3,3), (2,2,2), $u_3$] &    & \\
    \midrule
    9 & $^{*}x_4$=[(3,3,3), (2,2,2), $u_3$] &    & $x_{6,7}$ \\

    \bottomrule
    
\end{tabular}
    \caption{Trace of $\mathit{Open}$ and $\mathit{Sols}$ in each iteration (It.) of \textsf{NWRCA*}. We mark extracted node of each iteration with symbol $^{*}$. The third column shows changes on $\mathrm{G_{cl}^{Tr}}$ lists.}
    \label{tab:example}
\end{table}

\noindent
It.1: The search starts with expanding the initial node $x_0$. Since $x_0$ is the first expansion of $u_s$ and thus non-dominated, its truncated cost is stored in $\mathrm{G_{cl}^{Tr}}(u_s)$, as shown in the third column of Table~\ref{tab:example}.
Expansion of $x_0$ will add three new nodes ($x_1,x_2,x_3$) to $\mathit{Open}$, all showing valid resource estimates.\\
It.2: Among the nodes in $\mathit{Open}$, $x_1$ shows the smallest $f_1$-value and is expanded first. $x_1$ is non-dominated, and its expansion generates $x_4$ and $x_5$, associated with $u_3$ and $u_g$ respectively. 
We have ${\bf f}(x_4) = (3,3,3)$ and ${\bf f}(x_5) = (-1,1,4)$.
$x_5$ is pruned as its estimated resource usage is not within the requested resource limit ($f_3(x_5) > 3$).
$x_4$, however, is added to the $\mathit{Open}$ as it shows valid estimates.
\\
It.3: Both $x_2$ and $x_3$ show the same $f_1$-value. $x_3$ is extracted based on the LIFO strategy. $x_3$ is non-dominated, thus its truncated cost vector $(1,1)$ is stored in $\mathrm{G_{cl}^{Tr}}(u_3)$. $x_3$'s expansion generates the new node $x_6$. This new node shows valid resource usage and is added to $\mathit{Open}$.\\
It.4: $x_6$ is extracted with $u_g$. After updating $\mathrm{G_{cl}^{Tr}}(u_g)$ with the truncated cost vector $(2,2)$, our first solution node can be stored in $\mathit{Sols}$. $x_6$ does not need to be expanded, but it updates the primary upper bound $\overline{f_1} \gets 2$.\\
It.5: $x_2$, the second descendant node of $x_0$, is extracted.
$x_2$ is non-dominated, and its expansion generates two nodes $x_7$ and $x_8$, associated with $u_g$ and $u_3$, respectively. Both nodes show valid resource estimates and are added to $\mathit{Open}$.\\
It.6: Both $x_7$ and $x_8$ show the same $f_1$-value. $x_8$ is extracted based on the LIFO strategy. $x_8$ is non-dominated, because its truncated cost vector $(2,0)$ is not dominated by the other cost vector $(1,1)$. Thus, $\mathrm{G_{cl}^{Tr}}(u_3)$ can be updated. Expansion of $x_8$ generates $x_9$ with $u_g$. This new node is not dominated by the last expansion of $u_g$ and shows valid resource estimates. Thus, it will be added to $\mathit{Open}$.\\
It.7: $x_9$ is extracted with $u_g$ and is non-dominated. It will be added to $\mathit{Sols}$, but it cannot dominate $x_6$ (previous solution).\\
It.8: $x_7$ is extracted. $x_7$ is also a non-dominated solution node.
However, $x_{7}$'s truncated cost vector $(3,0)$ dominates and thus replaces the last vector in $\mathrm{G_{cl}^{Tr}}(u_g)$. $x_{7}$ also dominates the previous solution and thus replaces $x_9$ in $\mathit{Sols}$.\\
It.9: $x_4$ is extracted. Given the upper bound of primary cost updated in It.4, the search terminates due to $x_4$ surpassing the upper bound $\overline{f_1}$ with $f_1(x_4) = 3 > 2$.

The example above shows how \textsf{NWRCA*} processes nodes in the order of their $f_1$-value.
As we observed in It.7, \textsf{NWRCA*} may capture a dominated solution due to not breaking ties between nodes before extractions, but we discussed in It.8 how the search can refine the solution set in such circumstances to produce non-dominated optimal paths.

\section{Theoretical Results}
This section provides a formal proof for the correctness of constrained search of \textsf{NWRCA*}. We provide theoretical results on why the algorithm can solve RCSP instances with negative weights.
Throughout this section, we assume negative cycles have already been handled via Line~\ref{alg:high-level:cycle} of Algorithm~\ref{alg:high-level}, and thus our constrained A* search will not observe negative cycle on any $\mathit{start}$-$\mathit{goal}$ paths. 
Further, we assume that ${\bf h}$ is a perfect heuristic function, satisfying both consistency and admissibility requirements of constrained search with A* \cite{AhmadiTHK23_Networks}.

\noindent \textbf{Lemma 1\ }
Suppose \textsf{NWRCA*}'s search is led by smallest (possibly negative) $f_1$-values.
Let $x_i$ and $x_{i+1}$ be nodes extracted from $\mathit{Open}$ in two consecutive iterations of the search.
We have $f_1(x_i) \leq f_1(x_{i+1})$ if $h_1$ is consistent.

\noindent \textbf{Proof Sketch\ }
We distinguish two cases:
i) if $x_{i+1}$ was available in $\mathit{Open}$ at the time $x_i$ was extracted, the lemma is trivially true.
ii) otherwise, $x_{i+1}$ is the descendant node of $x_{i}$.
For the edge linking state $s(x_i)$ to its successor $s(x_{i+1})$, the consistency requirement of $h_1$ ensures $h_1(s(x_i)) \leq h_1(s(x_{i+1})) + \mathit{cost}_1(s(x_i), s(x_{i+1}))$.
Adding the cost $g_1(x_i)$ to both sides of the inequality yields $f_1(x_i) \leq f_1(x_{i+1})$. 
\hfill $\square$

\noindent \textbf{Corollary 1\ }
Let $(x_1,x_2,...,x_t)$ be the sequence of nodes extracted from $\mathit{Open}$. The (perfect) heuristic function ${\bf h}$ is consistent and admissible. Then, under the premises of Lemma~1, $i \leq j$ implies $f_1(x_i) \leq f_1(x_j)$, meaning $f_1$-values of extracted nodes are monotonically non-decreasing.

\noindent \textbf{Lemma 2\ }
Suppose $x_j$ is extracted after $x_i$ and $s(x_i)=s(x_j).$
$x_i$ weakly dominates $x_j$ if $\mathrm{Tr}({\bf g}(x_i)) \preceq \mathrm{Tr}({\bf g}(x_j))$.

\noindent \textbf{Proof Sketch\ }
$x_j$ is extracted after $x_i$, so we have $f_1(x_i) \leq f_1(x_j)$ according to Corollary~1.
Given $h_1(s(x_i)) = h_1(s(x_j))$, we obtain $g_1(x_i) \leq g_1(x_j)$.
The other condition $\mathrm{Tr}({\bf g}(x_i)) \preceq \mathrm{Tr}({\bf g}(x_j))$ means $g_2(x_i) \leq g_2(x_j), \dots, g_{d+1}(x_i) \leq g_{d+1}(x_j)$.
Thus, ${\bf g}(x_j)$ is no smaller than ${\bf g}(x_i)$ in all dimensions.
\hfill $\square$

\noindent \textbf{Lemma 3\ }
Dominated nodes cannot lead to any non-dominated $\mathit{cost}_1$-optimal solution path.

\noindent \textbf{Proof Sketch\ }
We prove this lemma by assuming the contrary, namely by claiming that dominated nodes can lead to a non-dominated $\mathit{cost}_1$-optimal solution path.
Let $x$ and $y$ be two nodes associated with the same state where $y$ is dominated by $x$.
Suppose that $\pi^*$ is a non-dominated $\mathit{cost}_1$-optimal $\mathit{start}$-$\mathit{goal}$ solution path via the dominated node $y$. 
Since $x$ dominates $y$, one can replace the subpath represented by $y$ with that of $x$ on $\pi^*$ to further reduce the $\mathit{\bf cost}$ of the optimum path represented by $y$ at least in one dimension (either primary cost or resources).
However, being able to reduce the $\mathit{\bf cost}$ of the established optimal solution path would either contradict our assumption on the optimality of the solution path $\pi^*$, or its non-dominance in the presence of (one or more) lower resources.
Therefore, we conclude that dominated nodes cannot form any non-dominated $\mathit{cost}_1$-optimal solution path. 
 \hfill $\square$

\noindent \textbf{Lemma 4\ }
Let $y$ be a node weakly dominated by $x$ and $s(x)=s(y)$.
If $y$'s expansion leads to a non-dominated $\mathit{cost}_1$-optimal solution path, $x$'s expansion will also lead to a non-dominated $\mathit{cost}_1$-optimal solution.

\noindent \textbf{Proof Sketch\ }
We prove this lemma by assuming the contrary, namely that $x$ cannot lead to any non-dominated $\mathit{cost}_1$-optimal solution path.
Since $y$ is weakly dominated by $x$, we have $g_1(x) \leq g_1(y),\dots,g_{d+1}(x) \leq g_{d+1}(y)$, meaning that $x$ offers a better cost at least in one dimension, or $\mathit{\bf cost}$ equal to the $\mathit{\bf cost}$ of $y$.
In such condition, one can replace the partial path represented by $y$ with that of $x$, and nominate a path lexicographically smaller than or equal to the optimal solution path via $y$, even through a (non-negative) cycle.
The existence of a better solution in the former case would contradict our assumption on the non-dominance or optimailty of the solution via node $y$.
The latter case means both paths are equal in terms of $\mathit{\bf cost}$ and thus should be either non-optimal or dominated, contradicting our assumption.
Therefore, node $x$ will definitely lead to a non-dominated $\mathit{cost}_1$-optimal solution path if node $y$ also leads to such path.
 \hfill $\square$

 \noindent \textbf{Lemma 5\ }
Node $x$ cannot lead to any optimal $\mathit{start}$-$\mathit{goal}$ solution path if $\mathit{\bf f}(x) \nleq \overline{\bf f}$.

\noindent \textbf{Proof Sketch\ }
$\mathit{\bf f}(x) \nleq \overline{\bf f}$ means either $f_1(x) > \overline{f_1}$ or $f_{k+1}(x) > R_k$ for at least one resource $k\in \{1 \dots d\}$.
Since ${\bf h}$ is admissible, the latter condition ensures that expanding $x$ towards $\mathit{goal}$ does not lead to a solution within the resource limits. 
The former also guarantees that the $x$'s expansion cannot lead to a solution with $\mathit{cost}_1$-value better than any of the current solutions in $\mathit{Sols}$ with the optimal cost of $\overline{f_1}$.
\hfill $\square$

\noindent \textbf{Theorem 1\ }
\textsf{NWRCA*} computes a set of non-dominated $\mathit{cost}_1$-optimal paths for negative-cycle free RCSP problems.

\noindent \textbf{Proof Sketch\ }
\textsf{NWRCA*} enumerates all partial paths from the $\mathit{start}$ state towards the $\mathit{goal}$ state in best-first order, in search of all optimal solutions.
The dominance pruning strategies utilised by \textsf{NWRCA*} (Lemmas~2) ensure that removal of (weakly) dominated nodes is safe, as they will not lead to non-dominated $\mathit{cost}_i$-optimal solution paths (Lemmas 3-4).
Nodes violating the upper bounds can also be pruned safely (Lemma~5).
Thus, we just need to show that the algorithm terminates with no weakly dominated solution in $\mathit{Sols}$.
\textsf{NWRCA*} prunes all weakly dominated nodes but captures all non-dominated nodes reaching the $\mathit{goal}$ state.
However, since it does not process nodes lexicography, some tentative solutions may later appear dominated.
Let $x$ be a new non-dominated solution extracted after solution $z$.
We must have $f_1(x) = f_1(z)$, otherwise the algorithm was terminated if $f_1(x) > \overline{f_1}=f_1(z)$.
In this case, a dominance check is performed to ensure $z$ is non-dominated, or remove $z$ from $\mathit{Sols}$ if it is deemed to be dominated by $x$ (in all resources), as scripted in lines~\ref{alg:nwrca:sol1}-\ref{alg:nwrca:sol2} of Algorithm~\ref{alg:nwrca}.
When the search surpasses $\overline{f_1}$, according to Corollary~1, expansion of remaining nodes in $\mathit{Open}$ cannot lead to solutions with a primary cost better than $\overline{f_1}$, and thus, the termination criterion is correct.  
Therefore, we conclude that \textsf{NWRCA*} terminates with returning a set of non-dominated $\mathit{cost}_1$-optimal solution paths in negative-cycle free RCSP problems.
 \hfill $\square$

\section{Empirical Analysis}
We compare our \textsf{NWRCA*} with the state-of-the-art RCSP approaches and evaluate them on 400 instances from \citeauthor{AhmadiTHK22_socs}~\shortcite{AhmadiTHK22_socs} over four maps from the 9th DIMACS Implementation Challenge \cite{dimacs9th}, with the largest map containing over 1~M nodes and 2.6~M edges. We have 100 instances per map.
Following the literature, we define the resource limit $\mathit{R_k}$ based on the tightness of the constraint $\delta$ as:
\begin{equation*}
\delta=\dfrac{R_{k}-h_{k+1}}{ub_{k+1}-h_{k+1}} \quad \text{for} \ \delta \in\{10\%,30\%,50\%,70\%,90\%\}  
\end{equation*}
where $h_{k+1}$ and $ub_{k+1}$ are, respectively, lower and upper bounds on $\mathit{cost}_{k+1}$ of $\mathit{start}$-$\mathit{goal}$ paths.
The upper bound $ub_{k+1}$ is set by the (non-constrained) $\mathit{cost}_1$-optimal path.
We choose the same $\delta$ for all resources $k \in \{1,\dots,d\}$.
We study scenarios with two and three resources ($d=2,3$).
The road networks in DIMACS maps, however, only provide two edge attributes: distance (our primary cost) and time.
Following \citeauthor{ren2023erca}~\shortcite{ren2023erca}, we extend the dimension and define the third cost of each edge as a function of its average (out)degree, whereas the fourth cost is set to one (denoting the depth/intersections).
Although our algorithm can handle negative costs and resources, we do not add negative attributes to the edges so we can also evaluate the existing methods. 

\begin{table}[t]
\centering
\small
\setlength{\tabcolsep}{4.5pt}
\begin{tabular}{|l | l |r | *{3}{r} | *{1}{r}|}
\toprule
     & & & \multicolumn{3}{c|}{Runtime(s)} & \multicolumn{1}{c|}{} \\ \cmidrule{4-6}
     Map & Algorithm & $|S|$ & \multicolumn{1}{c}{Min.} & \multicolumn{1}{c}{Mean} & \multicolumn{1}{c|}{Max.} & \multicolumn{1}{c|}{$\phi$}\\
  
    \midrule
\multicolumn{7}{|c|}{Instances with two resources} \\
\midrule

NY       & \textsf{NWRCA*} & 100     & \textbf{0.06}   & \textbf{1.35}  & \textbf{46.3}   & 1.00    \\
       &  \textsf{ERCA*}  & 99     & 1.79   & 95.22  & 3600.0   & 30.90    \\
       & \textsf{RCBDA*}  & 94    & 0.52   & 372.68  & 3600.0  & 115.28    \\
\midrule
BAY & \textsf{NWRCA*}  & 100 & \textbf{0.10}  & \textbf{2.99}    & \textbf{83.0}     & 1.00   \\
 & \textsf{ERCA*}    & 99  & 1.95 & 214.19  & 3600.0   & 28.59  \\
 & \textsf{RCBDA*}    & 92  & 0.61 & 433.32  & 3600.0   & 105.68 \\
\midrule
COL & \textsf{NWRCA*}  & 100 & \textbf{0.19} & \textbf{21.01}   & \textbf{1200.4} & 1.00   \\
 & \textsf{ERCA*}    & 92  & 2.66 & 583.31  & 3600.0   & 36.00   \\
 & \textsf{RCBDA*}    & 78  & 0.82 & 984.59  & 3600.0   & 89.16  \\
\midrule
FLA & \textsf{NWRCA*}  & 99  & \textbf{0.91} & \textbf{407.32}  & 3600.0   & 1.00   \\
 & \textsf{ERCA*}    & 41  & 8.26 & 2317.39 & 3600.0   & 36.17  \\
 & \textsf{RCBDA*}    & 32  & 3.61 & 2598.85 & 3600.0  & 60.42   \\
\midrule

\multicolumn{7}{|c|}{Instances with three resources} \\
    \midrule
NY  & \textsf{NWRCA*} & 100 & \textbf{0.08}  & \textbf{18.95}   & \textbf{662.3}  & 1.00      \\
    & \textsf{ERCA*}  & 90  & 2.31  & 568.57  & 3600.0   & 43.62  \\
    & \textsf{RCBDA*}   & 80  & 0.69  & 792.16  & 3600.0   & 72.47  \\
\midrule
BAY & \textsf{NWRCA*} & 100 & \textbf{0.14}  & \textbf{35.74}   & \textbf{1566.8} & 1.00      \\
    & \textsf{ERCA*}  & 88  & 2.69  & 607.86  & 3600.0   & 30.45  \\
    & \textsf{RCBDA*}   & 74  & 0.80   & 966.76  & 3600.0   & 49.95  \\
\midrule
COL & \textsf{NWRCA*} & 99  & \textbf{0.28}  & \textbf{118.26}  & 3600.0   & 1.00      \\
    & \textsf{ERCA*}  & 73  & 3.44  & 1136.46 & 3600.0   & 40.17  \\
    & \textsf{RCBDA*}   & 66  & 1.10   & 1471.62 & 3600.0   & 175.75 \\
\midrule
FLA & \textsf{NWRCA*} & 72  & \textbf{1.27}  & \textbf{1471.45} & 3600.0   & 1.00      \\
    & \textsf{ERCA*}  & 22  & 11.24 & 2877.43 & 3600.0   & 37.37  \\
    & \textsf{RCBDA*}   & 20  & 5.83  & 2937.15 & 3600.0   & 39.79  \\
    \bottomrule
\end{tabular}
\caption{Runtime statistics of the algorithms (in seconds) with $k=2,3$. $|S|$ is the number of solved cases (out of 100) within one hour (timeout), and $\phi$ shows the average slowdown factor of mutually solved cases with respect to the virtual best oracle. 
}
\label{table:results_3c}
\end{table}

\begin{figure}[t]
\centering
\includegraphics[width=1\columnwidth]{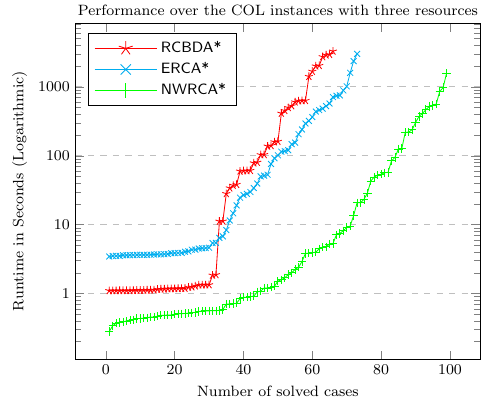}
\caption{Algorithms' performance over the instances of the COL map with $k=3$. Solved instances are sorted based on runtime.}
\label{fig:cactus_performance}
\end{figure}

\noindent
\textbf{Baseline algorithms:} We consider \textsf{RCBDA*} of \citeauthor{thomas2019exact}~\shortcite{thomas2019exact} and also the recent \textsf{ERCA*} method \cite{ren2023erca} as baselines. 
We do not consider the \textsf{BiPulse} algorithm \cite{cabrera2020exact} as it is already shown to be outperformed by \textsf{ERCA*} in \citeauthor{ren2023erca}~\shortcite{ren2023erca}.

\noindent
\textbf{Implementation:}
We implemented our \textsf{NWRCA*} algorithm in C++ and used the publicly available version of the \textsf{ERCA*} algorithm also implemented in C++.
For the \textsf{RCBDA*} algorithm, we were unable to obtain the original implementations. 
We, therefore, implemented the algorithm in C++ based on the descriptions provided in the original paper.
To navigate the search in \textsf{NWRCA*}, we use bucket-based queues (with the LIFO strategy) as they have been empirically shown to be more efficient than conventional binary heaps in constrained search with A* \cite{AhmadiTHK23_Networks}.
Note that the correctness of the \textsf{NWRCA*} algorithm does not depend on the lexicographical ordering of nodes in the priority queue.
All C++ code was compiled using the GCC7.5 compiler. 
We ran all experiments on a single core of an Intel Xeon-Platinum-8260 processor running at 2.5GHz and with 32GB of RAM, under the CentOs Linux 7 environment and a one-hour timeout.

Table~\ref{table:results_3c} compares the performance of the algorithms with two and three resources ($k=2,3$). 
We report for each set of experiments the number of solved cases ($|S|$ out of 100), the minimum, (arithmetic) mean and maximum runtime observed. 
The runtime includes the time each algorithm needs to compute lower bounds.
To compare the performance on each individual instance, we report in the last column the average slow-down factor (of runtime) over the mutually solved instances with respect to a virtual best oracle (denoted by $\phi$).
For every mutually solved instance, the virtual oracle is given the best runtime of the three algorithms, so $\phi$-value denotes how slow each algorithm is on average when compared with the virtual best oracle.
We observe from the results that \textsf{NWRCA*} significantly outperforms the baselines, and has consistently resulted in larger number of solved cases. 
For instances with two resources (three cost components), for example, \textsf{NWRCA*} has been able to solve 100\%\ of the cases for the NY, BAY, and COL maps within 20~minutes, whereas the second-best algorithm \textsf{ERCA*} is unable to solve 1-8 instances over the maps within the one-hour timeout. 
The same pattern is also seen in instances with $k=3$ and we observe far more cases solved with \textsf{NWRCA*}.
The outperformance is even more pronounced in the FLA map, where \textsf{NWRCA*} could solve 72\%\ of the cases, whereas \textsf{ERCA*} and \textsf{RCBDA*} could solve less than one-fourth of the cases.
Comparing the performance against the virtual best oracle, our detailed results highlight that \textsf{NWRCA*} performs best in all mutually solved instances. 
Further, we can see \textsf{ERCA*} and \textsf{RCBDA*} performing up to 43 and 175 times slower than \textsf{NWRCA*} across mutually solved instances on average.
Comparing \textsf{ERCA*} and \textsf{RCBDA*}, we can see \textsf{ERCA*} performs faster in all maps based on the average values.
We also report in Figure~\ref{fig:cactus_performance} the number of solved cases of the algorithms when sorted by runtime in the COL map with $k=3$.
We can find \textsf{NWRCA*} solving around 70\% of instances within 10 seconds, whereas the other algorithms need above 1000 seconds to reach the 70\% success rate. 
It also solves nearly half of the instances before \textsf{ERCA*} solves its easiest instance.
In summary, \textsf{NWRCA*} presents far faster performance compared to the other algorithms in terms of computation time.

\section{Conclusion}
This paper introduced \textsf{NWRCA*}, a fast solution to RCSP on the basis of A* search.
\textsf{NWRCA*} leverages the efficient pruning strategies introduced in recent constrained and multi-objective pathfinding technologies and introduces a new search framework that can efficiently solve large RCSP instances in limited time.
\textsf{NWRCA*} can deal with RCSP problems with both negative costs and negative resources as long as there is no negative cycle along any paths of the given ($\mathit{start}$,$\mathit{goal}$) pair.
The results of our experiments over a set of realistic RCSP instances demonstrate the effectiveness of \textsf{NWRCA*} in solving difficult RCSP instances, where it outperforms the state of the art by up to two orders of magnitude.

\section{Acknowledgements}
This research was supported by the Department of Climate Change, Energy, the Environment and Water under the International Clean Innovation Researcher Networks program grant number ICIRN000077.
Mahdi Jalili is supported by Australian Research Council through projects DP240100963, DP240100830, LP230100439 and IM240100042.

\bibliographystyle{named}
\bibliography{References.bib}

\end{document}